\documentclass[11pt,a4paper]{article}
\usepackage[hyperref]{naaclhlt2019}
\usepackage{times}
\usepackage{latexsym}
\usepackage{graphicx}
\usepackage{url}
\usepackage{amsmath}
\usepackage{color}
\usepackage{hyperref}
\usepackage{rotating}
\usepackage{multirow}
\usepackage{pbox}
\usepackage{float}
\usepackage{caption}
 \aclfinalcopy 

\title{WikiHow: A Large Scale Text Summarization Dataset}

\author{Mahnaz Koupaee \\
  University of California, Santa Barbara \\
  {\tt koupaee@cs.ucsb.edu} \\\And
  William Yang Wang \\
 University of California, Santa Barbara \\
  {\tt william@cs.ucsb.edu} \\}
\date{}

\begin{document}
\maketitle
\begin{abstract}
Sequence-to-sequence models have recently gained the state of the art performance in summarization. However, not too many large-scale high-quality datasets are available and almost all the available ones are mainly news articles with specific writing style. Moreover, abstractive human-style systems involving description of the content at a deeper level require data with higher levels of abstraction.
In this paper, we present WikiHow, a dataset of more than 230,000 article and summary pairs extracted and constructed from an online knowledge base written by different human authors. The articles span a wide range of topics and therefore represent high diversity styles. We evaluate the performance of the existing methods on WikiHow to present its challenges and set some baselines to further improve it. 
\end{abstract}

\section{Introduction}
Summarization as the process of generating a shorter version of a piece of text while preserving important context information is one of the most challenging NLP tasks. Sequence-to-sequence neural networks have recently obtained significant performance improvements on summarization~\cite{rush2015neural,chopra2016abstractive}. However, the existence of large-scale datasets is the key to success of these models. Moreover, the length of the articles and the diversity in their styles can create more complications. 

Almost all existing summarization datasets such as DUC \cite{harman2004effects}, Gigaword \cite{napoles2012annotated}, New York Times \cite{sandhaus2008new} and CNN/Daily Mail \cite{nallapati2016abstractive} consist of news articles. The news articles have their own specific styles and therefore the systems trained on only news may not be generalized well. On the other hand, the existing datasets may not be large enough (DUC) to train a sequence-to-sequence model, the summaries may be limited to only headlines (Gigaword), they may be more useful as an extractive summarization dataset (New York Times) and their abstraction level might be limited (CNN/Daily mail).

To overcome the issues of the existing datasets, we present a new large-scale dataset called WikiHow using the online WikiHow\footnote{\url{http://www.wikihow.com/}} knowledge base. It contains articles about various topics written in different styles making them different form existing news datasets.
Each article consists of multiple paragraphs and each paragraph starts with a sentence summarizing it. By merging the paragraphs to form the article and the paragraph outlines to form the summary, the resulting version of the dataset contains more than 200,000 long-sequence pairs. 
We then present two features to show how abstractive our dataset is. Finally, we analyze the performance of some of the existing extractive and abstractive systems on WikiHow as benchmarks for further studies.
The contribution of this work is three-fold:
\begin{itemize}
\item We introduce a large-scale, diverse dataset with various writing styles, convenient for long-sequence text summarization.
\item We introduce level of abstractedness and compression ratio metrics to show how abstractive the new dataset is.
\item We evaluate the performance of the existing systems on WikiHow to create benchmarks and understand the challenges better.
\end{itemize}



\section{Existing Datasets}
There are several datasets used to evaluate the summarization systems. We briefly describe the properties of these datasets as follows.

\noindent\textbf{DUC:} The Document Understanding Conference dataset \cite{harman2004effects} contains 500 news articles and their summaries capped at 75 bytes. The summaries are written by human authors and there exist more than one summary per article which is its major advantage over other existing datasets. 
The DUC dataset cannot be used for training models with large number of parameters and therefore is used along with other datasets \cite{rush2015neural,nallapati2017summarunner}. 
\begin{figure}[t]
  \centering    
  \includegraphics[width=0.37\textwidth]{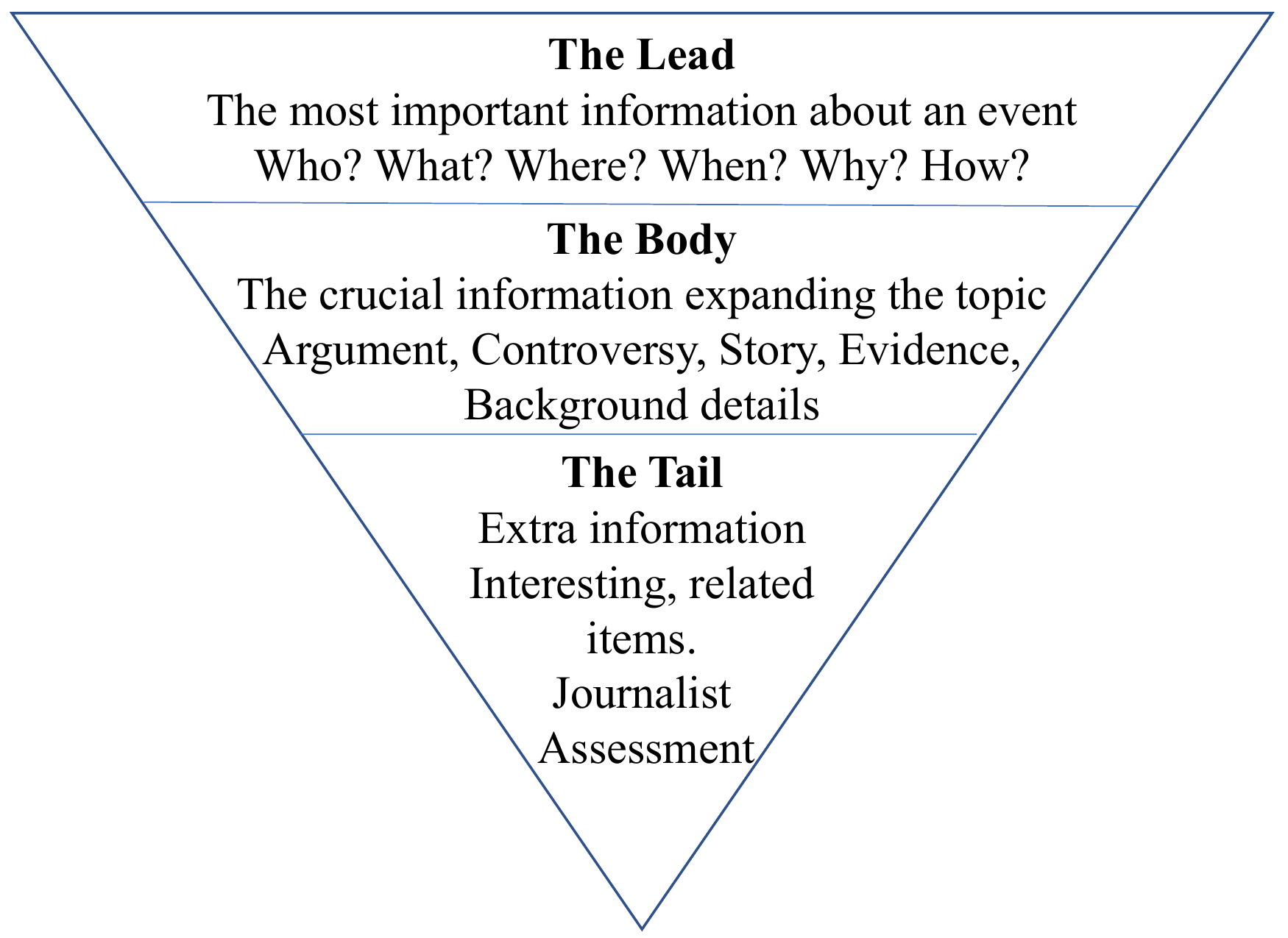}
  \caption{Inverted Pyramid writing style. The first few sentences of news articles contain the important information making Lead-3 baselines outperforming most of the systems.}
  \label{fig:pyramid}
  \vspace*{-2ex}
\end{figure}

\noindent\textbf{Gigaword:} Another collection of news articles used for summarization is Gigaword \cite{napoles2012annotated}. The original articles in the dataset do not have summaries paired with them. However, some prior work \cite{rush2015neural,chopra2016abstractive} used a subset of this dataset and constructed pairs of summaries by using the first line of the article and its headline, making the dataset suitable for short text summarization tasks.

\noindent\textbf{New York Times:} The New York Times (NYT) dataset \cite{sandhaus2008new} is a large collection of articles published between 1996 and 2007. While this dataset has been mainly used for extractive systems \cite{hong2014improving,durrett2016learning}, \citet{paulus2017deep} are the first to evaluate their abstractive system using NYT.

\noindent\textbf {CNN/Daily Mail:} This dataset mainly used in recent summarization papers ~\cite{nallapati2016abstractive,see2017get,nallapati2017summarunner} consists of online CNN and Daily Mail news articles and was originally developed for question/answering systems. The highlights associated with each article are concatenated to form the summary. Two versions of this dataset depending on the preprocessing exist. \citet{nallapati2017summarunner} has used the entity anonymization to create the anonymized version of the dataset while \citet{see2017get} replaced the anonymized entities with their actual values and create the non-anonymized version.

\noindent\textbf {NEWSROOM:} This corpus \cite{grusky2018newsroom} is the most recent large-scale dataset introduced for text summarization. It consists of diverse summaries combining abstractive and extractive strategies yet it is another news dataset and the average length of summaries are limited to $26.7$.
\section{WikiHow Dataset}
\begin{table}[t]
\footnotesize
\label{tbl:1}
\centering
\begin{tabular}{lcc}
\hline
Dataset Size&230,843\\
Average Article Length&579.8\\
Average Summary Length&62.1\\
Vocabulary Size&556,461\\
\hline
\end{tabular}
\caption{The WikiHow datasets statistics.}
 \vspace*{-2ex}
\end{table} 
The existing summarization datasets, consist of news articles. These articles are written by journalists and follow the journalistic style. The journalists usually follow the Inverted Pyramid style \cite{po2003news} (depicted in \hyperref[fig:pyramid]{Figure 1}) to prioritize and structure a text by starting with mentioning the most important, interesting or attention-grabbing elements of a story in the opening paragraphs and later adding details and any background information. This writing style might be the cause why lead-3 baselines (where the first three sentences are selected to form the summary) usually score higher compared to the existing summarization systems.  
We introduce a new dataset called WikiHow, obtained from WikiHow data dump. This dataset contains articles written by ordinary people, not journalists, describing the steps of doing a task throughout the text. Therefore, the Inverted Pyramid does not apply to it as all parts of the text can be of similar importance. 

\subsection{WikiHow Knowledge Base}
\begin{figure*}[t]
  \centering    
  \includegraphics[width=\textwidth]{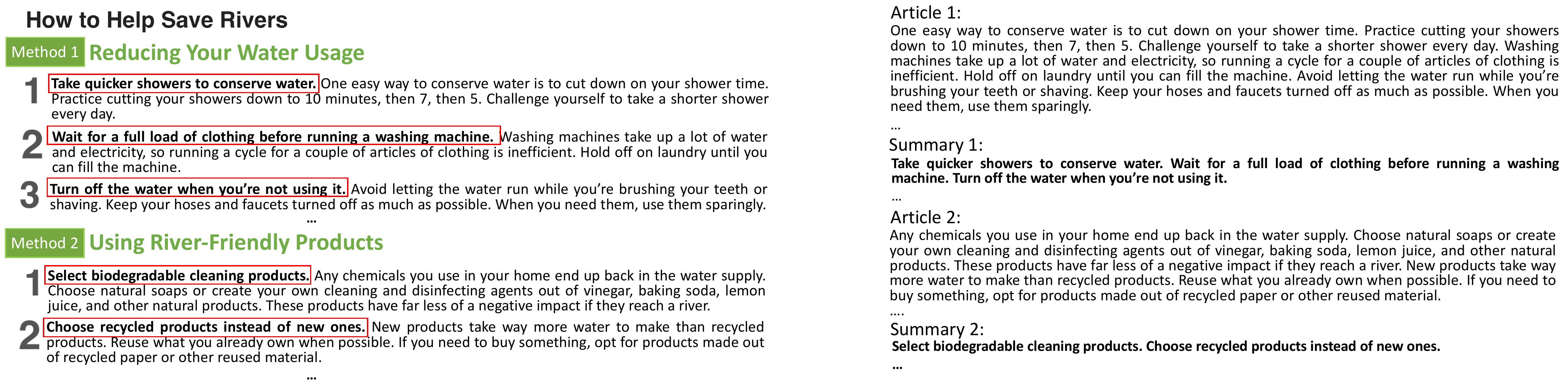}
  \caption{An example of our new dataset: WikiHow summary dataset, which includes +200K summaries. The bold lines summarizing the paragraph (shown in red boxes) are extracted and form the summary. The detailed descriptions of each step (except the bold lines) will form the article. Note that the articles and the summaries are truncated and the presented texts are not in their actual lengths.}
  \label{fig:wikihow}
\end{figure*}

The WikiHow knowledge base contains online articles describing a procedural task about various topics (from arts and entertainment to computers and electronics) with multiple methods or steps and new articles are added to it regularly. Each article consists of a title starting with \textbf{``How to''} and a short description of the article. There are two types of articles: the first type of articles describe single-method tasks in different steps, while the second type of articles represent multiple steps of different methods for a task. Each step description starts with a bold line summarizing that step and is followed by a more detailed explanation. A truncated example of a WikiHow article and how the data pairs are constructed is shown in \hyperref[fig:wikihow]{Figure 2}. 

\subsection{Data Extraction and Dataset Construction}
We made use of the python Scrapy \footnote{\url{https://scrapy.org/}} library to write a crawler to get the data from the WikiHow website. The articles classified into $20$ different categories, cover a wide range of topics. Our crawler was able to obtain $142,783$ unique articles (some containing more than one method) at the time of crawling (new articles are added regularly). 
To prepare the data for the summarization task, each method (if any) described in the article is considered as a separate article. To generate the reference summaries, bold lines representing the summary of the steps are extracted and concatenated. The remaining parts of the steps (the detailed descriptions) are also concatenated to form the source article. After this step, $230,843$ articles and reference summaries are generated. 
There are some articles with only the bold lines i.e. there is no more explanation for the steps, so they cannot be used for the summarization task. To filter out these articles, we used a size threshold so that pairs with summaries longer than the article size will be removed. The final dataset is made of $204,004$ articles and their summaries. 
The statistics of the dataset are shown in \hyperref[tbl:1]{Table 1}. The dataset is released to the public\footnote{\url{https://github.com/mahnazkoupaee/WikiHow-Dataset}}.

\section{WikiHow Properties}
The large scale of the WikiHow dataset by having more than $230,000$ pairs, and its average article and summary lengths makes it a better choice compared to DUC and Gigaword corpus.
We also define two metrics to represent the abstraction level of WikiHow by comparing it with CNN/Daily mail known as one of the most abstractive and common datasets in recent summarization papers \cite{nallapati2016abstractive,nallapati2017summarunner,see2017get,paulus2017deep}.
\begin{figure}
  \centering    
  \includegraphics[width=0.45\textwidth]{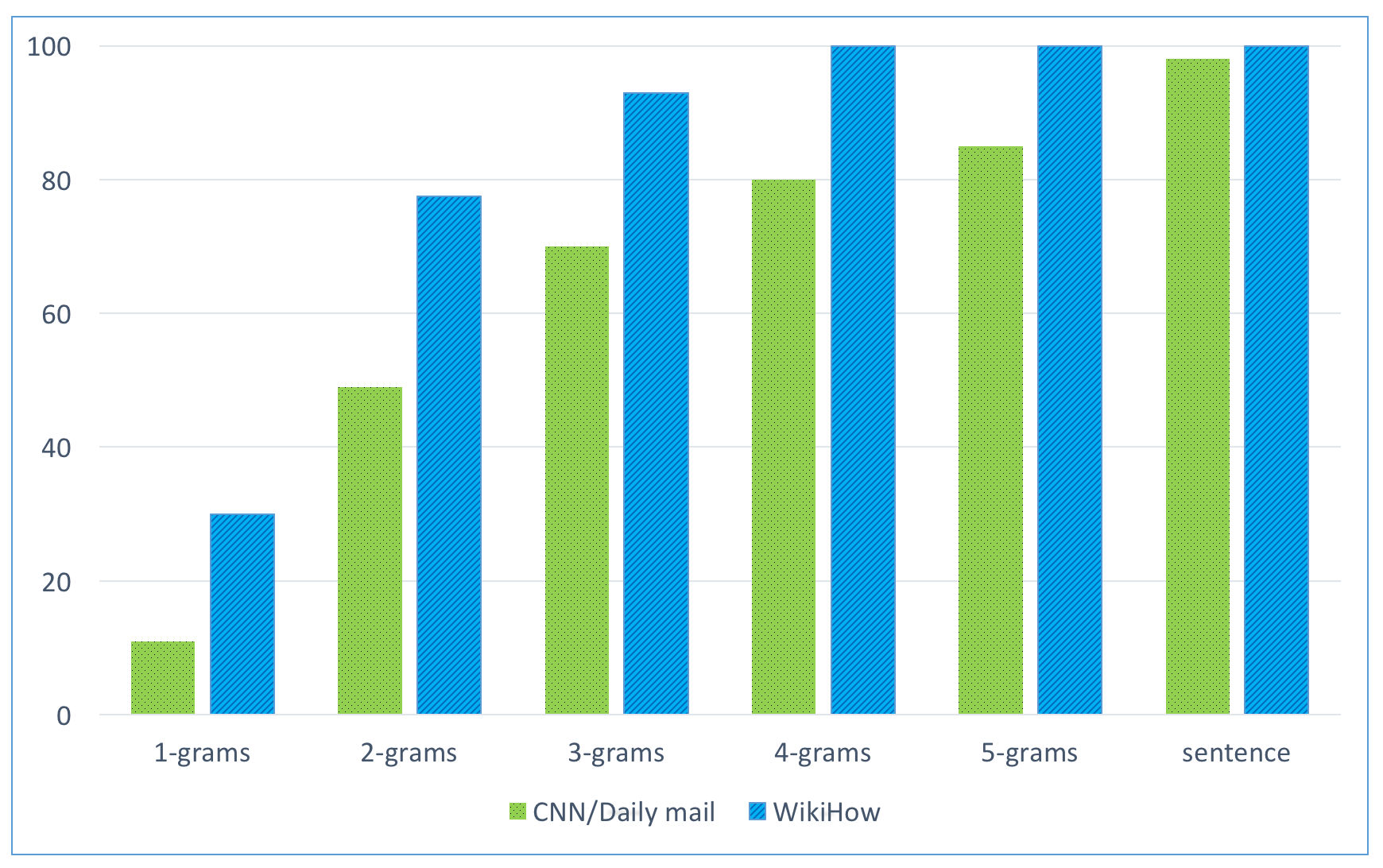}
  \caption{Uniqueness of n-grams in CNN/Daily mail and WikiHow datasets.}
  \label{fig:abstract}
  \vspace*{-2mm}
\end{figure}


\begin{table*}
\footnotesize
\label{results}
\centering
\begin{tabular}{|l|c|c|c|c|c|c|c|c|}
\hline
&\multicolumn{4}{|c|}{CNN/Daily Mail}&\multicolumn{4}{|c|}{WikiHow}\\
\cline{2-9}
\multicolumn{1}{|c|}{Model}&\multicolumn{3}{|c|}{ROUGE}&\multicolumn{1}{|c|}{METEOR}&\multicolumn{3}{|c|}{ROUGE}&\multicolumn{1}{|c|}{METEOR}\\
\cline{2-9}
&1&2&L&exact&1&2&L&exact\\
\hline
{TextRank} &35.23&13.90&31.48&18.03&27.53&7.4&20.00&\textbf{12.92}\\
{Seq-to-seq with attention} &31.33&11.81&28.83&12.03&22.04&6.27&20.87&10.06\\
{Pointer-generator} &36.44&15.66&33.42&15.35&27.30&9.10&25.65&9.70\\
{Pointer-generator + coverage} &39.53&17.28&36.38&17.32&\textbf{28.53}&\textbf{9.23}&\textbf{26.54}&10.56\\
\hline
{Lead-3 baseline} &\textbf{40.34}&\textbf{17.70}&\textbf{36.57}&\textbf{20.48}&26.00&7.24 &24.25&12.85\\
\hline
\end{tabular}
\caption{The ROUGE-F1 scores of different methods on non-anonymized version of CNN/Daily Mail dataset and WikiHow dataset. The ROUGE scores are given by the 95\% confidence interval of at most $\pm0.25$ in the official ROUGE script.}
\end{table*}

\subsection{Level of Abstractedness}
Abstractedness of the dataset is measured by calculating the unique n-grams in the reference summary which are not in the article. 
The comparison is shown in \hyperref[fig:abstract]{Figure 3}. Except for common unigrams, bi-grams and trigrams between the articles, and the summaries, no other common n-grams exist in the WikiHow pairs. The higher level of abstractedness creates new challenges for the summarization systems as they have to be more creative in generating more novel summaries.

\subsection{Compression Ratio}
We define compression ratio to characterize the summarization. We first calculate the average length of sentences for both the articles and the summaries. The compression ratio is then defined as the ratio between the average length of sentences and the average length of summaries. The higher the compression ratio, the more difficult the summarization task, as it needs to capture higher levels of abstraction and semantics. \hyperref[compression]{Table 3} shows the results for WikiHow and CNN/Daily Mail. The higher compression ratio of WikiHow shows the need for higher levels of abstraction.
\begin{table}
\footnotesize
\centering
\begin{tabular}{lcc}
&WikiHow&CNN/Daily Mail\\
\hline
Article Sentence Length&100.68&118.73\\
Summary Sentence Length&42.27&82.63\\
Compression Ratio&2.38&1.44\\
\hline
\end{tabular}
\caption{Compression ratio of WikiHow and CNN/Daily mail datasets. The represented article and summary lengths are the mean over all sentences.}
\label{compression}
\vspace*{-2mm}
\end{table}

\section{Experiments} 
We evaluate the performance of the WikiHow dataset using existing extractive and abstractive baselines. The systems used and the results generated for WikiHow and CNN/Daily mail are described in the following sections.
\subsection{Evaluated Systems}
\textbf{TextRank Extractive system:} An extractive summarization system \cite{mihalcea2004textrank,barrios2016variations} using a graph-based ranking method to select sentences from the article and form the summary.

\noindent\textbf{Sequence-to-sequence model with attention:} A baseline system applied by \citet{chopra2016abstractive,nallapati2016abstractive} to abstractive summarization task to generate summaries using the predefined vocabulary. This baseline is not able to handle Out of Vocabulary words (OOVs).

\noindent\textbf{Pointer-generator abstractive system:} A pointer-generator mechanism \cite{see2017get} allowing the model to freely switch between copying a word from the input sequence or generating a word form the predefined vocabulary. 

\noindent\textbf{Pointer-generator with coverage abstractive system:} The pointer-generator baseline with added coverage loss \cite{see2017get} to reduce the repetition in the final generated summary.

\noindent\textbf{Lead-3 baseline:} A baseline selecting the first three sentences of the article to form the summary. 
This baseline cannot be directly used for the WikiHow dataset as the first $3$ sentences of each article only describe a small portion of the whole article. We created the Lead-3 baseline by extracting the first sentence of each paragraph and concatenated them to create the summary.


\subsection{Results}
To study the performance of the evaluated systems, we used the Pyrouge package\footnote{\url{pypi.python.org/pypi/pyrouge/0.1.3}} to report the F1 score for ROUGE-1, ROUGE-2 and ROUGE-L \cite{lin2004rouge} and the METEOR \cite{banerjee2005meteor}\footnote{\url{www.cs.cmu.edu/~alavie/METEOR}} both based on the exact matches and on inclusion of stem, paraphrasing and synonyms (s/p/s) to evaluate the methods . \hyperref[results]{Table 2} represents the results of multiple baselines on both the CNN/Daily Mail (the well-known, most common abstractive summarization dataset) and also the proposed WikiHow dataset. As it can be seen, the summarization systems perform a lot better on CNN/Daily mail compared to the WikiHow dataset with lead-3 outperforming other baselines due to the news inverted pyramid writing style described earlier. On the other hand, the poor performance of lead-3 on WikiHow shows the different writing styles in its articles. Moreover, all baselines perform about $10$ ROUGE scores better on the CNN/Daily mail compared to the WikiHow. This difference suggests new features and aspects inherent in the new dataset which can be used to further improve the summarization systems. 
\section{Conclusion}
We present WikiHow, a new large-scale summarization dataset consisting of diverse articles 
form WikiHow knowledge base. The WikiHow features discussed in the paper can create new challenges to the summarization systems. We hope that the new dataset can attract researchers attention as a choice to evaluate their systems. 
\bibliography{emnlp2018.bib,all.bib,mahnaz.bib}
\bibliographystyle{acl_natbib_nourl.bst}



\end{document}